\documentclass{article} 
\usepackage{iclr2024_conference,times}

\usepackage{amsmath,amsfonts,bm}

\def\eqref#1{equation~\ref{#1}}

\def\1{\bm{1}}

\DeclareMathAlphabet{\mathsfit}{\encodingdefault}{\sfdefault}{m}{sl}
\SetMathAlphabet{\mathsfit}{bold}{\encodingdefault}{\sfdefault}{bx}{n}

\usepackage{hyperref}
\usepackage{wrapfig}
\usepackage{graphicx}
\usepackage{graphics}
\usepackage{booktabs}
\usepackage{subfigure}
\usepackage{amssymb}
\usepackage{array}
\usepackage{amsthm}
\usepackage{color}
\usepackage{enumitem}
\usepackage{multirow}
\usepackage{amsmath}
\usepackage{booktabs}
\usepackage{algorithmic}
\usepackage{dsfont}
\usepackage{url}
\usepackage[skip=0pt]{caption}

\usepackage{etoolbox}
\newcommand{\zerodisplayskips}{%
  \setlength{\abovedisplayskip}{2pt}%
  \setlength{\belowdisplayskip}{2pt}%
  \setlength{\abovedisplayshortskip}{2pt}%
  \setlength{\belowdisplayshortskip}{2pt}}
\appto{\normalsize}{\zerodisplayskips}
\appto{\small}{\zerodisplayskips}
\appto{\footnotesize}{\zerodisplayskips}

\newcommand{\myparatight}[1]{\smallskip\noindent{\bf {#1}:}~}
\newcommand{\name}{\text{DRMGuard}}
\title{Defending Deep Regression Models against Backdoor Attacks}

\author{Lingyu Du$^1$, Yupei Liu$^2$, Jinyuan Jia$^2$, Guohao Lan$^1$ \\
$^1$Delft University of Technology, $^2$The Pennsylvania State University\\
}

\iclrfinalcopy 
\begin{document}

\maketitle

\begin{abstract}

Deep regression models are used in a wide variety of safety-critical applications, 
but are vulnerable to backdoor attacks. Although many defenses have been proposed for classification models, they are ineffective as they do not consider the uniqueness of regression models. First, the outputs of regression models are continuous values instead of discretized labels. Thus, the potential infected target of a backdoored regression model has infinite possibilities, which makes it impossible to be determined by existing defenses. Second, the backdoor behavior of backdoored deep regression models is triggered by the activation values of all the neurons in the feature space, which makes it difficult to be detected and mitigated using existing defenses. To resolve these problems, we propose {\name}, the first defense to identify if a deep regression model in the image domain is backdoored or not. {\name} formulates the optimization problem for reverse engineering based on the unique output-space and feature-space characteristics of backdoored deep regression models. We conduct extensive evaluations on two regression tasks and four datasets. The results show that {\name} can consistently defend against various backdoor attacks. We also generalize four state-of-the-art defenses designed for classifiers to regression models, and compare {\name} with them. The results show that {\name} significantly outperforms all those defenses.

\end{abstract}

\section{Introduction}

Regression techniques are widely used to solve tasks where the goal is to predict continuous values. Unsurprisingly, similar to their classification counterparts, regression techniques have been revolutionized with deep learning and have achieved the state-of-the-art result in many real-world applications. Examples such as gaze estimation~\citep{zhang2017mpiigaze,zhang19_pami}, head pose estimation~\citep{Borghi_2017_CVPR,kuhnke2019deep}, 
and facial landmark detection~\citep{sun2013deep,wu2019facial}, among many others~\citep{lathuiliere2019comprehensive}. Unfortunately, deep regression models (DRM) inherited the vulnerabilities of deep neural networks~\citep{badnet,liutrojaning2018,nguyen2021wanet,nguyen2020input,turner2019label} and did not escape from the threat of backdoor attacks. Existing work~\citep{sun2022backdoor} shows that an attacker can inject a backdoor trigger into a DRM such that it outputs an attacker-chosen target vector for any input stamped with an attacker-chosen backdoor trigger, while its predictions for clean inputs are unaffected. Therefore, given the wide adoption of DRM in many safety-critical applications such as driver attention monitoring \citep{bmwForbes,smartEye}
, navigation of autonomous vehicles~\citep{Zeisl_2015_ICCV}, and pedestrian attention monitoring~\citep{raza2018appearance,schulz2012video}, backdoor attacks raise severe safety concerns about the trustworthiness and robustness of DRMs. 

Existing solutions to defend deep classification model (DCM) against backdoor attacks can be divided into 
\emph{data-level}~\citep{chen2018detecting,guo2023scale} and \emph{model-level defenses}~\citep{liu2019abs,wang2019neural,wang2022rethinking}. Data-level defenses aim to detect backdoored training or testing data, while model-level defenses aim to detect a potentially backdoored model and unlearn the backdoor behaviors.
As we will discuss in Section~\ref{background_and_related_work}, our work focuses on model-level defenses, as they are more realistic and do not assume the defender has access to the backdoored training or testing data. 

However, existing model-level defenses~\citep{wang2022rethinking,wang2019neural,wu2021anp} are designed for DCMs. Our experiments show that they are ineffective when generalized and applied to DRMs that are considered in this work. There are two underlying causes. First, distinct from DCMs~\citep{Wang_2017_CVPR} for which the output space is discretized into a few class labels, the output space of DRMs is continuous. Thus, it is infeasible (if not impossible) to enumerate and analyze all the potential target vectors using existing defenses designed for DCMs to determine the infected target \citep{wang2019neural} or the compromised neurons \citep{liu2019abs}. 
Second, different from DCMs that adopt $\arg\max$ to obtain the final output, DRMs do not need $\arg\max$. 
This makes the backdoor behavior of the backdoored DRMs different from that of the backdoored DCMs. Specifically, for a backdoored DCM, the backdoor behavior is often triggered by the activation values of several neurons in the feature space \citep{wang2022rethinking}, whereas for a backdoored DRM, it is triggered by the activation values of all the neurons, which makes it harder to be detected or mitigated. 

\myparatight{Our work} In this paper, we propose {\name}, the first framework to detect backdoored DRMs in the image domain. {\name} is applied to a DRM to reverse engineer a potential trigger function, based on which we make the decision on whether the model has been injected a backdoor or not. A major challenge to reverse engineering of the potential trigger function in the regression domain is that the output is defined in the continuous space. To address this challenge, we formulate the reverse engineering as an optimization problem, which is based on both output-space and feature-space characteristics of the backdoored DRMs that are observed in this paper.


To demonstrate the effectiveness of {\name}, we consider two regression tasks, 
and conduct extensive experiments on four datasets for state-of-the-art backdoor attacks. Our experimental results suggest that {\name} is consistently effective in defending both input-independent attacks, e.g., BadNets~\citep{badnet}, and input-aware attacks, e.g., Input-aware dynamic attack \citep{nguyen2020input}. 
Furthermore, we adapt four state-of-the-art backdoor defenses, i.e., Neural Cleanse \citep{wang2019neural}, FeatureRE~\citep{ wang2022rethinking}, ANP~\citep{wu2021anp}, and Fine-pruning~\citep{liu2018fine}, designed for classifiers to regression models and compare {\name} with them. The results demonstrate that {\name} outperforms all of them by a large margin.

\section{Background and Related Work}
\label{background_and_related_work}

\myparatight{Backdoor Attacks} 
Many backdoor attacks~\citep{
chen2017targeted,badnet,liutrojaning2018,9747582, wang2022invisible, Wang_2022_CVPR, yao2019latent, Zhao_2022_CVPR} have been proposed for deep neural networks. They showed that an attacker can inject a backdoor into a classifier and make it output an attacker-chosen target class for any input embedded with an attacker-chosen backdoor trigger. 
Depending on whether the attacker uses the same backdoor trigger for different testing inputs, we categorize existing attacks into \emph{input-independent attacks}~\citep{chen2017targeted, badnet, liutrojaning2018, turner2019label, yao2019latent} and \emph{input-aware attacks}~\citep{9869920, Li_2021_ICCV, nguyen2021wanet,nguyen2020input, 9797338}. 
For instance, \citet{badnet} proposed an input-independent backdoor attack by using a fixed pattern, e.g., a white patch, as the backdoor trigger. 
Recently, researchers proposed to use input-aware techniques, such as the warping process \citep{nguyen2021wanet} and generative models \citep{nguyen2020input}, to generate dynamic triggers varying from input to input.
When extending those attacks to DRMs~\citep{sun2022backdoor}, an attacker can inject a backdoor and make the model output a fixed vector (called \emph{target vector}) for any testing input with the backdoor trigger. 
Lastly, backdoor attacks were also studied for graph neural networks~\citep{xi2021graphbackdoor,zhang2021backdoorgraph} and natural language processing~\citep{shen2021transfertoall}. They are out of the scope of this paper as we focus on attacks in the image domain.

\myparatight{Existing Defenses} 
We categorize existing defenses against backdoor attacks into \emph{data-level defenses}~\citep{doan2020februus, gao2019strip,ma2022beatrix} and \emph{model-level defenses}~\citep{liu2022unlearning,liu2019abs,wu2021anp,9296553,zeng2022ibau,zheng2022lips}. Data-level defenses 
detect whether a training example or a testing input is backdoored or not. 
They usually have 
two major limitations: 1) training data detection defenses \citep{chen2018detecting} are not applicable for a given model that is already backdoored; and 2) testing input detection defenses~\citep{doan2020februus} need to inspect each testing input at the running time and incur extra computation cost, and thus are undesired for latency-critical applications, e.g., gaze estimation \citep{Zhang2020ETHXGaze}. Therefore, we focus on model-level defense in this work. 

Model-level defenses detect whether a given model is backdoored or not, and state-of-the-art methods~\citep{Guan_2022_CVPR,qiao2019defending,wang2019neural,wang2022rethinking,9296553} are based on trigger reverse engineering. 
Specifically, they view each class as a potential target class and reverse engineer a backdoor trigger for it. Given the reverse-engineered backdoor triggers, they use statistical techniques to determine whether the classification model is backdoored or not. However, existing solutions are mainly designed for classification tasks that have categorical output. As we will show in this paper, they cannot be applied to DRMs. We note that \citet{li2021backdoor} also studied backdoor defense for DRMs, but they only considered a specific attack designed in their paper where the inputs of the regression model are low-dimensional vectors, i.e., five dimensional vectors. {By contrast, we consider DRMs in the high-dimensional image domain.}

\section{Design of \name}
\label{sec:method}                                     

\subsection{Threat Model}

\myparatight{Deep regression model}
A deep regression model (DRM) is a deep neural network that maps an input to a  vector, i.e., $f: \mathcal{X} \mapsto \mathcal{Y}$, where $\mathcal{X}\subset \mathbb{R}^{N_w\times N_h \times N_c}$ represents the input space with width $N_w$, height $N_h$, and channel $N_c$; and $\mathcal{Y} \in \mathbb{R}^d$ represents the $d$-dimensional output space. Given a training dataset $\mathcal{D}_{tr}$ that contains a set of training examples, we define the following loss $\frac{1}{|\mathcal{D}_{tr}|}\sum_{(x,y) \in \mathcal{D}_{tr}}\ell(f(x),y)$, where $(x,y)$ is a training example in 
$\mathcal{D}_{tr}$ and $\ell$ is the loss function for the regression task (e.g., $\ell_2$ loss) to update the parameters of $f$. 

\myparatight{Backdoor attacks}
We consider existing backdoor attacks for classification models~\citep{badnet, nguyen2021wanet, nguyen2020input, turner2019label}, and adapt them to DRMs. Specifically, given a training dataset $\mathcal{D}_{tr}$, an attacker can add backdoor triggers to the training samples in $\mathcal{D}_{tr}$, and change their ground-truth annotations to an attacker-chosen vector, $y_T\in \mathcal{Y}$, known as the \emph{target vector}. The attacker can manipulate the training process. 
The backdoored DRM performs well on benign inputs, but outputs the target vector $y_T$ when the backdoor trigger is present in the input. Formally, we define the backdoor attack for DRMs as:
\begin{equation}
    f(x)=y, \; f(\mathcal{A}(x))=y_T,
    \label{backdooredDRM}
\end{equation}
where $f$ is the backdoored DRM, {$ x\in \mathcal{X}$} is the benign input; $y \in \mathcal{Y}$ is the benign ground-truth annotation; and $\mathcal{A}$ is the trigger function that constructs the poisoned input from the benign input.  

\myparatight{Evaluation metric for backdoor attacks} Given a set of poisoned inputs, we define attack error (AE) as the average regression error calculated from the output vectors and the target vector over all the poisoned inputs, to evaluate the performance of backdoor attacks on DRMs. AE can be regarded as the counterpart to the attack success rate for backdoor attacks on classification models.

\myparatight{Assumptions and goals of the defender} The defense goal is to identify if a DRM has been backdoored or not. Following existing defenses for backdoor attacks~\citep{liu2019abs, wang2019neural, wang2022rethinking}, we assume the defender can access the trained DRM and a small benign dataset $\mathcal{D}_{be}$ with correct annotations.

\subsection{Overview of {\name}}

\begin{wrapfigure}{r}{6.7cm}
\vspace{-14pt}
	\centering

\includegraphics[scale=0.29]{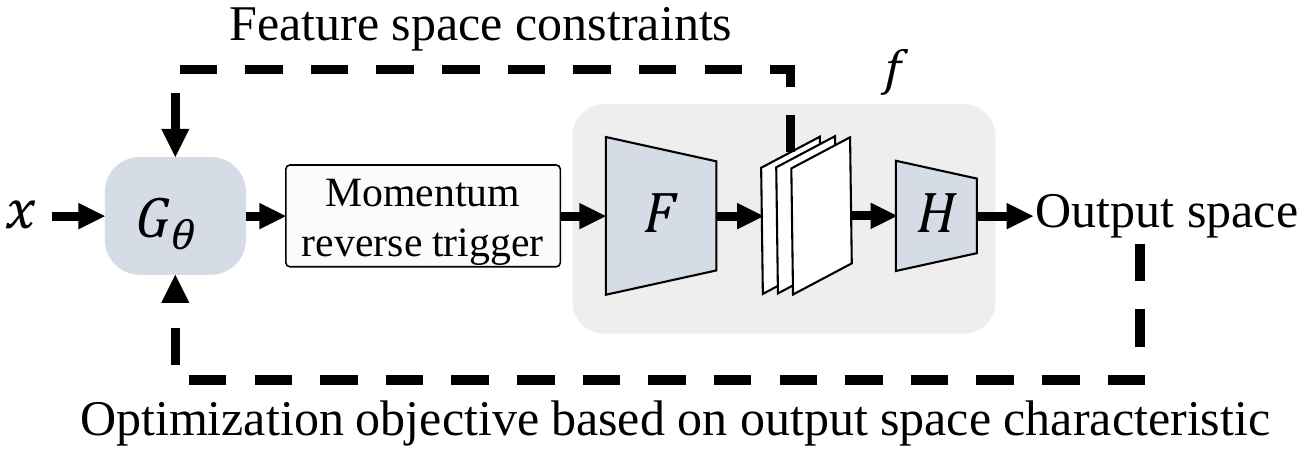}
	\caption{Overview of {\name}.
 }	
	\label{fig:overview}
	\vspace{-0.1in}
\end{wrapfigure}

We propose {\name} to identify if a DRM has been backdoored by reverse engineering the trigger function $\mathcal{A}$. {Figure \ref{fig:overview} shows the overview of {\name}.} We use a generative model $G_{\theta}$ to model $\mathcal{A}$. This allows us to model the trigger function for both input-independent and input-aware attacks. For a given DRM $f$ under examination, we split it into two submodels. Specifically, we first use the submodel $F$ to map the original input $x$ to the feature space $F(x) \in \mathbb{R}^m$, {i.e., the output space of the last but second layer of $f$.} 
Then, we use the submodel $H$ to map the intermediate feature from the feature space to the final output space. This allows us to investigate the characteristics of the backdoored DRMs in both feature space and output space, based on which we formulate the optimization problem for reverse engineering. Moreover, we propose a strategy called \emph{momentum reverse trigger} to reverse high-quality triggers.

\subsection{Observations and Intuitions for Backdoored Deep Regression Model}

Reverse engineering is performed by solving an optimization problem with constraints that are designed based on observations in the input \citep{wang2019neural} or the feature space \citep{wang2022rethinking}. Existing work \citep{wang2022rethinking} shows that by using feature-space constraints, one can reverse both input-independent trigger \citep{badnet} and input-aware trigger \citep{nguyen2021wanet, nguyen2020input} for a backdoored deep classification model (DCM). Following this trend, we consider feature-space constraints when designing the reverse engineering for DRM. Below, we first discuss the difference in the feature space between backdoored DCM and backdoored DRM. Then, through theoretical analysis and experiments, we introduce the key observation for backdoored DRM.

\myparatight{Difference between backdoored DCM and DRM} A key observation for backdoored DCMs is that the backdoor behavior is represented by the activation values of several neurons in the feature space~\citep{liu2019abs, wang2022rethinking}. Specifically, when a trigger is present in the input, the activation values of the affected neurons will drop into a certain range, making the backdoored DCM output the attacker-chosen target class regardless of the activation values of the other neurons. 
This is because, after applying a series of operations to the feature vector, a backdoored DCM utilizes $\arg\max$ to obtain the final classification output. 
As long as the activation values of the affected neurons can make the target class have the highest probability, the influence of the other neurons on the final classification output will be eliminated by $\arg\max$. 
By contrast, the final regression output of a backdoored DRM is obtained by applying linear transformation (or followed by an activation function) to the feature vector without using $\arg\max$. Thus, the activation value of each neuron in the feature space contributes to the final output. This difference inspires us to take all the neurons into consideration when searching for the feature-space characteristics of backdoored DRMs, rather than looking at a few specific neurons only. 

\myparatight{Theoretical analysis and metrics} We use $\{h_i\}_{i=1}^N$ and $\{h_i^p\}_{i=1}^N$ to denote the feature vectors extracted from a set of $N$ benign inputs $\{x_i\}_{i=1}^N$ and a set of poisoned inputs $\{\mathcal{A}(x_i)\}_{i=1}^N$, respectively, where $h_i=F(x_i)\in \mathbb{R}^m$ and $h_i^p=F(\mathcal{A}(x_i))\in\mathbb{R}^m$. 
We use $y_{i,j}$ and $y_{i,j}^p$ to denote the $j$th component of the output vector $y_i=H(h_i)\in \mathbb{R}^d$ and $y_i^p=H(h_i^p)\in \mathbb{R}^d$. $y_{i,j}^p$ is calculated by: 
\begin{equation}\small
    y_{i,j}^p = \Omega(w_j \cdot h_i^p + b_j) = \Omega(\|w_j\|_2\|h_i^p\|_2\cos{\alpha_{i,j}^p}+b_j),
    \label{eq:DRMoutputInner}
\end{equation}
where $\Omega(\cdot)$ is the activation function; $w_j \in \mathbb{R}^{m}$ and $b_j \in R$ are the weights vector and the bias of $H$ for the $j$th component of the output vector, respectively; $\alpha_{i,j}^p$ is the angle between $h_i^p$ and $w_j$. 
Based on Equation~\ref{backdooredDRM}, we have $y_{1,j}^p\approx y_{2,j}^p\approx\cdots\approx y_{N,j}^p$ if $f$ is backdoored, which means $\sigma^2(\{y_{i,j}^p\}_{i=1}^N)$ is a small positive value, where $\sigma^2(\cdot)$ is the variance function. As shown in Equation \ref{eq:DRMoutputInner}, the value of $\sigma^2(\{y_{i,j}^p\}_{i=1}^N)$ is influenced only by $\|h_i^p\|_2$ and $\alpha_{i,j}^p$, as $\|w_j\|_2$ and $b_j$ are constant for a given DRM. Moreover, when $f$ is backdoored, $\sum_{j=1}^{d}{\sigma^2(\{y_{i,j}^p\}_{i=1}^N)}/d$ is a small positive value and influenced by $\|h_i^p\|_2$ and $\alpha_i^p$, where $\alpha_i^p=\{\alpha_{i,1}^p,...,\alpha_{i,d}^p\}\in\mathbb{R}^d$. We use $\alpha_{i,j}$ to denote the angle between $h_i$ and $w_j$, and define $\alpha_i$ as $\alpha_i=\{\alpha_{i,1},...,\alpha_{i,d}\}\in\mathbb{R}^d$.

To further investigate how $\|h_i^p\|_2$ and $\alpha_{i}^p$ influence $\sum_{j=1}^{d}{\sigma^2(\{y_{i,j}^p\}_{i=1}^N)}/d$, we introduce \textbf{the ratio of norm variance (RNV)} and \textbf{the ratio of angle variance (RAV)}, as two feature-space metrics: \begin{gather}\small
\text{RNV}=\sigma^2(\{\|h_i^p\|_2\}_{i=1}^N)/\sigma^2(\{\|h_i\|_2\}_{i=1}^N) \; \textrm{and} \; 
\text{RAV}=\frac{1}{d}\sum_{j=1}^{d}{\sigma^2(\{\alpha_{i, j}^p\}_{i=1}^N)/\sigma^2(\{\alpha_{i, j}\}_{i=1}^N)}.
\end{gather}
Specifically, RNV compares the dispersion of $\{\|h_i^p\|_2\}_{i=1}^N$ and $\{\|h_i\|_2\}_{i=1}^N$, while RAV compares the dispersion of $\{\alpha_{i}^p\}_{i=1}^N$ and $\{\alpha_{i}\}_{i=1}^N$. $\text{RNV}\ll1$ indicates that when triggers are present in the inputs, the feature vectors extracted by $F$ 
have similar norms. $\text{RAV}\ll1$ means that the variance of angles between $\{h_i^p\}_{i=1}^N$ and $w_j$ are much smaller than that between $\{h_i\}_{i=1}^N$ and $w_j$ for $j=1,...,d$.

\begin{figure}[]
	\centering
	\subfigure[Backdoored DRMs trained on MPIIFaceGaze under different backdoor attacks $(d=2)$.]{\includegraphics[scale=0.51]{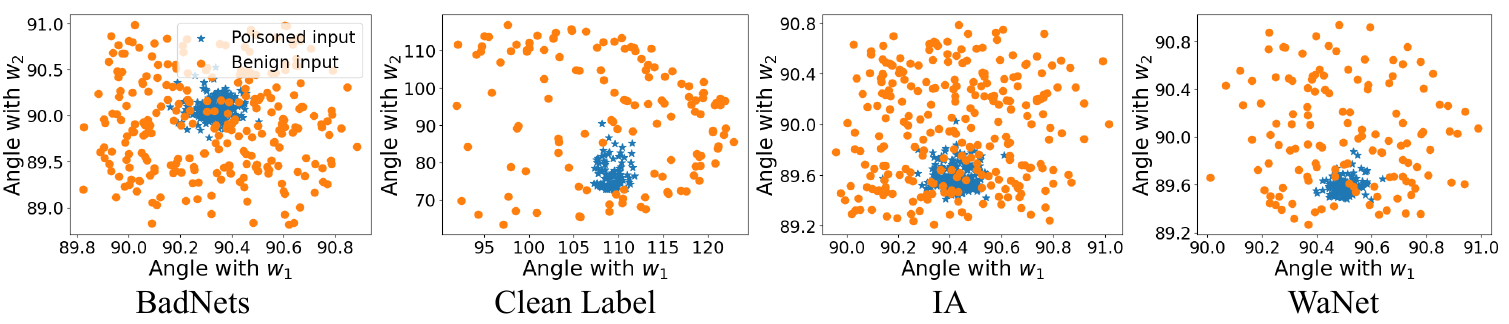}} 
	\subfigure[Backdoored DRMs trained on Biwi Kinect under different backdoor attacks $(d=3)$.]{\includegraphics[scale=0.51]{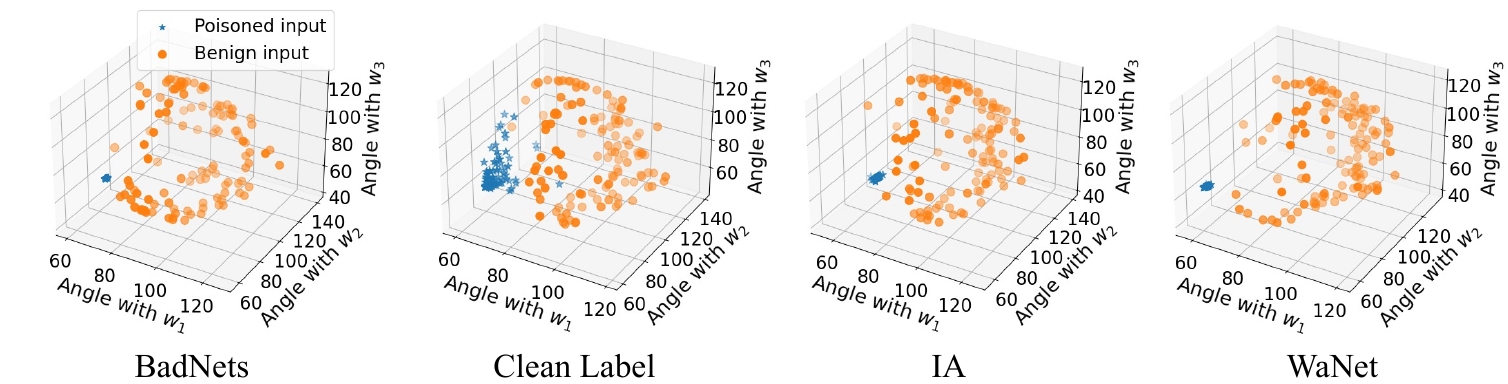}} 

\caption{The plots of $\{\alpha_{i}^p\}_{i=1}^N$ and $\{\alpha_{i}\}_{i=1}^N$ (in degree) for backdoored DRMs trained on (a) MPIIFaceGaze dataset and (b) Biwi Kinect dataset. 
The spread of the data points shows that the angles of the poisoned inputs are highly concentrated, while the angles of the benign inputs are scattered.}	
	\label{fig:key_observation}
	\vspace{-0.2in}
\end{figure}

\myparatight{Observations}
We use four backdoor attacks, i.e., BadNets \citep{badnet}, Input-aware dynamic attack (IA) \citep{nguyen2020input}, WaNet \citep{nguyen2021wanet}, and Clean Label \citep{turner2019label}, to train backdoored DRMs on MPIIFaceGaze dataset \citep{Zhang_2017_CVPR_Workshops} 
and Biwi Kinect dataset \citep{fanelli_IJCV}. 
Table \ref{Tab_observations} shows the RNV and the RAV of the backdoored DRMs that are trained by different backdoor attacks on the two datasets. 
The key observation is that RAV is significantly smaller than $0.1$ in all the examined scenarios. To further explore this observation, the scatter plots in Figure~\ref{fig:key_observation} visualize $\{\alpha_{i}^p\}_{i=1}^N$ and $\{\alpha_{i}\}_{i=1}^N$ in all the examined cases. We can see that the angles of the poisoned inputs are highly concentrated, while the angles of the benign inputs are scattered, meaning that $\sigma^2(\{\alpha_{i,j}^p\}_{i=1}^N) \ll \sigma^2(\{\alpha_{i,j}\}_{i=1}^N)$ for $j=1,\cdots,d$. We summarize this observation in the feature space as follows. 

\textbf{Key observation for backdoored DRMs in the feature space:} \textit{Consider a set of benign inputs $\{x_i\}_{i=1}^N$ and a 
set of poisoned inputs $\{\mathcal{A}(x_i)\}_{i=1}^N$. Let $W = \{w_1, w_2, ..., w_d\}^{\mathsf{T}} \in \mathbb{R}^{m\times d}$ be the weights matrix of $H$, where $w_j \in \mathbb{R}^{m}$. Then, we have the following observation: 
\begin{equation} \small
    \sigma^2\left(\big\{\mathcal{B}(F(\mathcal{A}(x_i)), w_j)\big\}_{i=1}^N\right) \ll \sigma^2\left(\big\{\mathcal{B}(F(x_i), w_j)\big\}_{i=1}^N\right) \;\; \textrm{for} \;\; j=1,2,...,d,
    \label{eq:keyObservation}
\end{equation} 
where $\mathcal{B}(v_1,v_2)=\arccos{(v_1 \cdot v_2)/(\|v_1\|_2\|v_2\|_2)}$; 
and $\mathcal{A}$ is the trigger function.}

\begin{wraptable}[9]{b}{6.5cm}
 \vspace{-12pt}
\centering
\caption{The RNV and RAV for four attacks on two datasets. In all the examined cases, {RAV} is significantly smaller than 0.1. 
}
\resizebox{\linewidth}{!}{
\begin{tabular}{ccccc}
\hline
\\[-1.5ex] 
\multirow{2}{*}{Attack} & \multicolumn{2}{c}{MPIIFaceGaze} & \multicolumn{2}{c}{Biwi Kinect} \\ \cline{2-5}\\[-1.5ex] 
                        & RAV             & RNV            & RAV            & RNV            \\ \hline\\[-1.5ex] 
BadNets                 & 0.0433          & 1.4499         & 0.0002         & 0.0012         \\
IA                      & 0.0489          & 2.5714         & 0.0015         & 0.0046         \\
Clean Label             & 0.0328          & 0.0428         & 0.0803         & 0.0341         \\
WaNet                   & 0.0311          & 0.8528         & 0.0003         & 0.0015         \\ \hline
\end{tabular}}
\label{Tab_observations}
\end{wraptable}

Lastly, Table \ref{Tab_observations} shows that 
$\text{RNV}$ is close to or greater than 0.1 for backdoored DRMs trained by BadNets, IA, and WaNet on MPIIFaceGaze, while for the other examined cases, $\text{RNV}$ is significantly smaller than 0.1. 
This is because, as shown in Figure \ref{fig:key_observation}, when $\text{RNV}\geq0.1$, 
each component of $\alpha_{i}^p$, i.e., $\alpha_{i,j}^p$ for $j=1,\cdots,d$, is almost 90$^\circ$, for $i=1,\cdots,N$, meaning that $\cos{\alpha_{i,j}^p}$ is almost zero. Thus, $y_{i,j}^p$ is insensitive to the change of $\|h_i^p\|_2$, which allows RNV to be similar to or even larger than 0.1 but still maintain a low $\sigma^2(\{y_{i,j}^p\}_{i=1}^N)$. 

\subsection{Methodology}
\myparatight{Reverse engineering for DRM}
One major challenge for reverse engineering for DRM is that the target vector $y_T$ is defined in the continuous output space. As a result, it is impossible to enumerate and analyze all the potential target vectors 
using existing reverse engineering methods designed for DCM, which treat each class as a potential target class and reverse the trigger for it~\citep{wang2019neural, wang2022rethinking}. To resolve this challenge, we propose to reverse engineer $\mathcal{A}$ by minimizing: $\sum_{j=1}^d{\sigma^2\left(\{f_j(G_{\theta}(x_i))\}_{i=1}^N\right)/d}$,
where $f_j(G_{\theta}(x_i))$ is the $j$th component of $f(G_{\theta}(x_i))\in \mathbb{R}^d$. This is intuitive, as for a backdoored $f$ the term $\frac{1}{d}\sum_{j=1}^d{\sigma^2\left(\{y_{i,j}^p\}_{i=1}^N\right)}$ will be a small positive value. 
Thus, 
we can search for the target vector in the continuous output space by learning a trigger function $\mathcal{A}$, modeled by $G_{\theta}$, that can mislead $f$ to map different inputs to 
the target vector without enumerating all the potential target vectors.
Moreover, based on the key observation of backdoored DRMs defined in Inequation~\ref{eq:keyObservation}, we 
introduce the feature-space regularization term in the optimization. Formally, we define the optimization problem for the reverse engineering for DRM as:
\begin{equation}
    \small
    \theta^{*}=\min_{\theta} \frac{\lambda_1}{d}\sum_{j=1}^d{\sigma^2\left(\{f_j\left(x'_i\right)\}_{i=1}^N\right)} + \frac{1}{N} \sum_{i=1}^N\|x'_i-x_i\|_1+\lambda_2 r_f,
    \label{opt:REforDRM}
\end{equation}

where $r_f = \frac{1}{d}\sum_{j=1}^{d}{{\sigma^2(\left\{\mathcal{B}(F(x'_i), w_j)\right\}_{i=1}^N) }\big/{\sigma^2(\left\{\mathcal{B}(F(x_i), w_j)\right\}_{i=1}^N) }}$; $x'_i=G_{\theta}(x_i)$; and $\lambda_1$ and $\lambda_2$ are the weights for the first and third objectives, respectively.
The first objective in the optimization problem aims to reverse engineer the poisoned inputs $\{G_{\theta}(x_i)\}_{i=1}^N$ that lead to the same output vector, regardless of their actual contents. 
The second objective is the input-space regularization term~\citep{wang2019neural, wang2022rethinking} that ensures the transformed input $G_{\theta}(x_i)$ is similar to the benign input $x_i$. 
The final objective is the feature-space regularization term, which can lead to a much lower variance for $\{\hat{\alpha}_{i,j}^p\}_{i=1}^N$ than that for $\{a_{i,j}\}_{i=1}^N$, where $\hat{\alpha}_{i,j}^p$ is the angle between $F(G_{\theta}(x_i))$ and $w_j$.

\myparatight{Momentum reverse trigger}
Directly solving the optimization problem defined in Equation~\ref{opt:REforDRM} can result in a sub-optimal solution (illustrated in Appendix \ref{App:momentum}): the algorithm focuses on the important regions in the input image, to which the DRM pay more attention, and directly adds perturbations to these regions to \emph{destroy} the task-related features. 
To avoid such a trivial solution, we propose the \emph{momentum reverse trigger} to assign different weights to different regions to balance the attention of the DRM on the image, such that the algorithm can pay attention to all the image pixels and search for the trigger that are injected into both important and unimportant regions. Details about the momentum reverse trigger are given in Appendix \ref{App:momentum}. After introducing the momentum reverse trigger into Equation \ref{opt:REforDRM}, we use $\mathcal{OPT}$-{\name} to denote the final optimization problem. 


\myparatight{Backdoor identification}
By solving the optimization problem $\mathcal{OPT}$-{\name}, we can obtain the perturbation $\|x'_i-x_i\|$ that transforms input $x_i$ to the potential target vector. We observe that the amount of perturbation required to transform the input to the potential target vector for a backdoored DRM is significantly smaller than that for a benign DRM. Based on this observation, {\name} introduces the metric $\small
    \mathcal{I}(f)=\mathds{1}(\frac{1}{N}\sum_{i=1}^{N}{\|x'_i-x_i\|_1}, \epsilon \|\hat{x}\|_1),
$ to identify if a given deep regression model is backdoored or not, where $\|\hat{x}\|_1$ is the input image that has the maximum $L1$ norm in {the benign dataset $\mathcal{D}_{be}$}; $\epsilon$ is a constant; and $\mathds{1}$ is the indicator function that returns $1$ (backdoored DRM) if $\frac{1}{N}\sum_{i=1}^{N}{\|x'_i-x_i\|_1}< \epsilon \|\hat{x}\|_1$ and $0$ (benign DRM) otherwise. We set $\epsilon=0.03$.

\myparatight{Backdoor mitigation}
Once a given DRM is identified as a backdoored DRM, {\name} uses the reversed trigger function $G_{\theta}$ and the available benign dataset $\mathcal{D}_{be}$ to generate a reversed poisoned dataset $\mathcal{D}_{rp}$ with the original correct annotations. Then, {\name} fine-tunes the given backdoored DRM by using $\mathcal{D}_{be}$ and $\mathcal{D}_{rp}$ to unlearn the backdoor behavior.

\section{Evaluation}
\vspace{-3mm}


\subsection{Experimental Setup}


\myparatight{Regression tasks} 
We consider two regression tasks, i.e., gaze estimation and head pose estimation. 
Gaze estimation tracks 
where the subject is looking at, and plays a key role in a series of safety-critical applications, such as user authentication \citep{eberz201928,katsini2020role} and driver distraction detection~\citep{smartEye}. 
Similarly, head pose estimation has also been used in many safety-related applications, such as the driver assistance system \citep{jha2016analyzing,murphy2007head} and pedestrian attention monitoring~\citep{schulz2012video}. 

\myparatight{Datasets} 
We consider four benchmark datasets, i.e., MPIIFaceGaze \citep{zhang19_pami}, ColumbiaGaze \citep{CAVE_0324}, Biwi Kinect \citep{fanelli_IJCV}, and Pandora \citep{Borghi_2017_CVPR}. 
For each dataset, we randomly select $80\%$ and $10\%$ of the images from the dataset to form the training dataset $\mathcal{D}_{tr}$ and the benign dataset $\mathcal{D}_{be}$, respectively. We use the remainder as the testing set $\mathcal{D}_{te}$ to evaluate the performance of backdoor mitigation. 
Details of datasets can be found in Appendix~\ref{App:datasets}.


\myparatight{Backdoor attacks}We consider four state-of-the-art backdoor attacks, including two input-independent attacks, i.e., BadNets~\citep{badnet} and Clean Label~\citep{turner2019label}, 
and two input-aware attacks, i.e., Iuput-aware dynamic attack (IA)~\citep{nguyen2020input} and WaNet~\citep{nguyen2021wanet}. We detail how to adapt these backdoor attacks to DRMs and the effectiveness of them on DRMs in Appendix~\ref{App:attack}. 

\myparatight{Compared defenses}
For backdoor identification, we compare {\name} with two state-of-the-art methods, i.e., Neural Cleanse (NC)~\citep{wang2019neural} and FeatureRE~\citep{wang2022rethinking}.
Since the output space of DRM is continuous, we generalize them from classification models to regression models by taking the potential target vector $y_t$ as the optimization variable. We provide details of the generalizations in Appendix \ref{App:baseline_defenses}. For backdoor mitigation, we compare {\name} with two state-of-the-art methods, i.e., ANP~\citep{wu2021anp} and Fine-pruning~\citep{liu2018fine}. 

\myparatight{Evaluation metrics} Following existing work~\citep{wang2022rethinking}, we use 
the identification accuracy as the performance metric. In detail, given a set of DRMs including benign and backdoored DRMs, the identification accuracy is defined as the percentage of correctly classified DRMs over all examined DRMs. We also report the number of True Positives (TP), i.e., correctly identified backdoored DRMs, False Positives (FP), i.e., benign DRMs recognized as backdoored DRMs, False Negatives (FN), i.e., backdoored DRMs identified as benign DRMs, and True Negatives (TN), i.e., correctly recognized benign DRMs. Moreover, we use ROC-AUC score to compare backdoor identification performance between {\name}, NC, and FeatureRE, after obtaining the average perturbations on $\mathcal{D}_{be}$ for benign DRMs and backdoored DRMs.  

To evaluate the performance of {\name} on backdoor mitigation, we generate a poisoned dataset $\mathcal{PD}_{te}$ by applying trigger function to all the images in $\mathcal{D}_{te}$. We define defending attack error (DAE) as the average regression error calculated from the output vectors and the correct annotations over all the images in $\mathcal{PD}_{te}$. Details of regression errors for the two examined regression tasks are given in Appendix \ref{App:datasets}. We use DAE and AE on $\mathcal{PD}_{te}$ as the evaluation metrics for backdoor mitigation.

\myparatight{Defense settings}
Unless otherwise mentioned, we set $\lambda_1=20$ and $\lambda_2=800$ for gaze estimation, and set $\lambda_1=10$ and $\lambda_2=100$ for head pose estimation, given task difference. We use ResNet18 \citep{He_2016_CVPR} (without the dense layer) to implement $F$, and a dense layer without activation function to implement $H$. 
{We consider gaze estimation task with MPIIFaceGaze dataset and the state-of-the-art input-aware attack WaNet.}

\begin{table}[]
\vspace{-10pt}
\begin{minipage}[c]{0.48\textwidth}
\centering
\caption{Backdoor identification performance of {\name} on MPIIFaceGaze for different attacks. {\name} can defend various attacks.
}
\resizebox{.9\linewidth}{!}{
\begin{tabular}{cccccc}
\hline
\\[-2ex]
       Attack                  & TP  & FP  & FN  & TN  & Acc  \\ \hline\\[-1.5ex]
       BadNets                 & 10    & 1    & 0    &9    &  95\%    \\
       IA                      & 10    & 1    & 0    &9   &95\%   \\
       Clean Label             & 10    & 1    & 0    &9    &  95\%     \\
       WaNet                   & 10    & 1    & 0    &9   & 95\%      \\\hline

\end{tabular}}
\label{Tab_evaluation_results_MPII}
\end{minipage}
\quad
\hfill
\begin{minipage}[c]{0.48\textwidth}
\centering
\caption{Backdoor identification performance of {\name} on different datasets for WaNet. {\name} is effective on various 
datasets.
}
\resizebox{.95\linewidth}{!}{
\begin{tabular}{cccccc}
\hline
\\[-2ex]

       Dataset                       & TP  & FP  & FN  & TN  & Acc  \\ \hline\\[-1.5ex]
      MPIIFaceGaze            & 10    &   1  &  0   &  9   & 95\%  \\
       ColumbiaGaze            & 8    & 4    & 2    & 6    &  70\%    \\
       Biwi Kinect             &   10  &   0  & 0    &10   &100\%   \\
       Pandora                 & 7    &   0  &  3   &  10   &  85\%    \\ \hline

\end{tabular}}
\label{Tab_evaluation_results_different_datasets}
\end{minipage}
\vspace{-8pt}
\end{table}

\subsection{Evaluation Results on Backdoor Identification}

\myparatight{{\name} is effective for backdoor identification} 
We conduct three experiments to evaluate the backdoor identification performance. 
{First, we evaluate the performance of {\name} in identifying backdoored DRMs trained by different attacks.} Specifically, for each of the four backdoor attacks, i.e., BadNets, Clean Label, WaNet, and IA, we train ten benign DRMs and ten backdoored DRMs on MPIIFaceGaze dataset. 
{The results are shown in Table \ref{Tab_evaluation_results_MPII}, which indicate that {\name} can identify backdoored DRMs trained by both input-independent and input-aware attacks, 
at an average accuracy of $95\%$.} Moreover, 
we visualize the estimation of the target vector during the training process and the reversed trigger in the Appendix \ref{App:target_vector} and \ref{App:reversed_trigger}, respectively.

Second, 
we examine the backdoor identification capability of {\name} on different regression tasks and datasets, i.e., MPIIFaceGaze, ColumbiaGaze, Biwi Kinect, and Pandora. Specifically, we train ten benign DRMs and ten backdoored DRMs using WaNet for each dataset. The results are shown in Table \ref{Tab_evaluation_results_different_datasets}. The average identification accuracy of {\name} on different datasets is $87.5\%$, which demonstrates the effectiveness of {\name} on various regression tasks and datasets. 

Finally, we consider the scenario where the DRM is backdoored by multiple trigger functions with different target vectors. We report the attacking details and evaluation results in Appendix \ref{App:multi_target_attack}. In brief, the results show that our method is effective on identifying DRMs with multiple backdoors.

\begin{table}[t]
\begin{minipage}[c]{0.51\textwidth}
\centering
\caption{ROC-AUC scores of different methods on MPIIFaceGaze for different attacks. {\name} significantly outperforms NC and FeatureRE.
}
\resizebox{.915\linewidth}{!}{
\begin{tabular}{cccc}
\hline\\[-2ex]
Attack     & NC    & FeatureRE   & DRMGuard \\ \hline\\[-2ex]
BadNets     & 0.270    & 0.730         & \textbf{1.000}    \\
IA          & 0.300    & 0.560         & \textbf{1.000}    \\
WaNet       & 0.940    & 0.560         & \textbf{1.000}    \\
Clean Label & 0.005    & 0.545         & \textbf{1.000}    \\ 
All attacks & 0.379    & 0.599         & \textbf{1.000}    \\ \hline
\end{tabular}}
\label{Tab_AUC_MPII}
\end{minipage}
\quad
\hfill
\begin{minipage}[c]{0.45\textwidth}
\centering
\caption{Performance of backdoor mitigation for different attacks. {\name} can mitigate backdoor behaviors for various attacks.
}

\label{Tab:backdoor_mitigation_diff_attack}
\resizebox{.92\linewidth}{!}{
\begin{tabular}{ccccc}
\hline
\\[-2ex]
\multirow{2}{*}{Attack} & \multicolumn{2}{c}{Undefended} & \multicolumn{2}{c}{\name} \\ \cline{2-5} \\[-2ex]
                        & AE             & DAE           & AE           & DAE           \\ \hline\\[-2ex]
BadNets                 & 3.25           & 14.85         & 17.21        & 3.59         \\ 
IA                      & 3.19           & 14.40          & 15.69          & 3.50         \\ 
Clean Label             & 0.72           & 15.43          & 16.42         & 2.51         \\ 
WaNet                   & 1.31           & 15.90          & 15.36        & 3.29        \\ \hline
\end{tabular}}
\end{minipage}
\vspace{-0.15in}
\end{table}

\myparatight{{\name} outperforms state-of-the-art defenses}
Table \ref{Tab_AUC_MPII} shows the ROC-AUC scores of {\name}, NC, and FeatureRE for four backdoor attacks. We also report the scores when applying the four backdoor attacks simultaneously. 
As shown, the ROC-AUC score of {\name} is $1.000$ in all the examined cases, which is significantly higher than that of NC and FeatureRE. Besides, we notice that FeatureRE fails to find a trigger function that enables the backdoored DRM to map different inputs to similar output vectors, which confirms our analysis that the feature-space characteristic for backdoored DCM~\citep{wang2022rethinking} does not hold for backdoored DRM. 

\subsection{Evaluation Results on Backdoor Mitigation}

\begin{wraptable}[8]{b}{5.cm}
\vspace{-13pt}
\centering
\caption{Performance of different methods on backdoor mitigation. {\name} outperforms baselines.
}
\resizebox{0.8\linewidth}{!}{%
\begin{tabular}{ccc}
\hline\\[-2ex]
Method       & AE    & DAE      \\ \hline \\[-2ex]
Undefended   & 1.31  & 15.90  \\
DRMGuard     & \textbf{15.36} & \textbf{3.29}   \\
Fine-tuning  & 13.96  & 4.32  \\
Fine-pruning & 6.68  & 16.82 \\
ANP          & 4.92  & 13.13  \\ \hline
\end{tabular}%
}
\label{Tab_mitigation_comparsion}
\end{wraptable}

We train backdoored DRMs using BadNets, Clean Label, IA, and WaNet on MPIIFaceGaze. Table \ref{Tab:backdoor_mitigation_diff_attack} shows AE and DAE of the undefended and mitigated backdoored DRMs, which indicate that {\name} can mitigate backdoor behaviors for various attacks. Specifically, {\name} can significantly increase AE and decrease DAE for all the attacks, which indicates that the output vectors of DRMs are far away from the target vector and close to the correct annotations after backdoor mitigation, even though triggers are injected in the inputs.


We compare {\name} with ANP and Fine-pruning on backdoor mitigation. We also consider a baseline, i.e., Fine-tuning, which directly uses the benign dataset $\mathcal{D}_{be}$ to fine tune the backdoored DRM. We report AE and DAE for different methods after backdoor mitigation in Table \ref{Tab_mitigation_comparsion}. The AE for {\name} is significantly larger than that for other methods, while the DAE for {\name} is much smaller than that for other methods,
which shows the superiority of {\name} on backdoor mitigation. Moreover, Fine-pruning and ANP are built upon the feature-space characteristics of backdoored DCM and perform terribly on backdoored DRMs. This also confirms our analysis that the feature-space characteristics of backdoored DRMs are different with that of backdoored DCM.

\subsection{Ablation Studies}

\myparatight{Impact of weights and the size of benign dataset}
To investigate the impact of $\lambda_1$ and $\lambda_2$ in Equation \ref{opt:REforDRM} on the performance of backdoor identification, we vary $\lambda_1$ and $\lambda_2$ from $10$ to $30$ and from $600$ to $800$, respectively. Moreover, 
we study the impact of the size of $\mathcal{D}_{be}$ on the identification performance by changing the ratio $p$ of benign dataset to the original whole dataset from $5\%$ to $15\%$. 

We report the results in Table \ref{Tab:abliation_study_parameters}. We observe that the performance of {\name} is insensitive to $\lambda_1$, as the identification accuracy is almost stable with different $\lambda_1$. However, {\name} is sensitive to $\lambda_2$ and the identification accuracy increases with $\lambda_2$. 
This observation proves that the proposed feature-space regularization term is important for the identification of backdoored DRMs. 
We also observe that as $p$ decreases from $15\%$ to $5\%$, the identification accuracy and the number of TN decrease, while the number of TP remains stable.
This is because, compared to a large $p$, it is easier to find a small amount of perturbation that can lead to the backdoor behavior on a small $p$ for benign models. 
However, the identification accuracy is still $90\%$ even when $p=5\%$.

\myparatight{Impact of feature-space regularization term (FSRT)} 
We remove FSRT from $\mathcal{OPT}$-{\name} and show the results in Table \ref{Tab:ablation_study}, which indicate that all the DRMs are classified as backdoored DRMs.
We further observe that without the FSRT, {\name} cannot find a trigger function that can map different inputs to similar output vectors. As a result, {\name} solves the optimization problem by focusing on minimizing the distance between the poisoned and benign images, and returns a small amount of perturbations, which leads to the misclassification 
of backdoored DRMs.

\myparatight{Impact of momentum reverse trigger (MTR)} 
We remove the MTR from $\mathcal{OPT}$-{\name} and report the identification results in Table \ref{Tab:ablation_study}. As shown, all the benign DRMs are classified as backdoored DRMs. This is because, without the MTR, {\name} will find a small amount of perturbation and add it to the eye regions to destroy gaze-related features and fool the benign models.
In this way, the poisoned images reversed from different images are transformed by $f$ to similar output vectors, and {\name} fails to correctly recognize benign DRMs. 

\begin{table}[]
\scriptsize

\centering
\caption{Ablation study on the impact of different value of $\lambda_1$, $\lambda_2$ and $p$.}
\resizebox{.85\linewidth}{!}{%
\begin{tabular}{cccccccccccc}
\hline\\[-2ex]
\multirow{2}{*}{Metric} & \multicolumn{3}{c}{Different $\lambda_1$} &  & \multicolumn{3}{c}{ Different $\lambda_2$} & & \multicolumn{3}{c}{Different $p$} \\ \cline{2-4} \cline{6-8} \cline{10-12} \\[-2ex]
                        & 10      & 20     & 30     &  & 600     & 800    & 1000  &  & $5\%$     & $10\%$    & $15\%$  \\ \hline\\[-2ex]
TP                      & 10      & 10     & 10     &  & 10      & 10     & 10   &  & 10       & 10         & 10   \\
FP                      & 0       & 1      & 1      &  & 5       & 1      & 0    &  & 2         & 1         & 0   \\
FN                      & 0       & 0      & 0      &  & 0       & 0      & 0    &  & 0         & 0          & 0   \\
TN                      & 10       & 9      & 9      &  & 5       & 9      & 10    &  & 8        & 9         & 10   \\
Acc                     & 100\%    & 95\%   & 95\%   &  & 75\%    & 95\%  & 100\% &  & 90\%     & 95\%      & 100\%  \\ \hline
\end{tabular}
\label{Tab:abliation_study_parameters}%
}
\vspace{-0.15in}
\end{table}

\begin{table}[]
\scriptsize
\hspace*{\fill}
\begin{minipage}[c]{0.51\textwidth}

\centering
\caption{Ablation study on FSRT and MTR.}
\label{Tab:ablation_study}
\resizebox{0.92\linewidth}{!}{%
\begin{tabular}{cccccc}
\hline
\\[-2ex]
Method   & TP & FP & FN & TN & Acc  \\ \hline\\[-2ex]
w/o FSRT  & 10 & 10 & 0  & 0  & 50\% \\ 
w/o MTR  & 10 & 10 & 0  & 0  & 50\%\\ \hline
\end{tabular}%
}
\end{minipage}
\quad
\hfill
\begin{minipage}[c]{0.46\textwidth}
\centering
\caption{Evaluation results on adaptive attack.}
\label{Tab:adaptive_attack}
\resizebox{0.79\linewidth}{!}{%
\begin{tabular}{llll}
\hline\\[-2ex]
Attack   & AE   & DAE   & Acc  \\ \hline\\[-2ex]
WaNet    & 1.51 & 15.99 & 95\% \\
Adaptive &  5.71    & 15.01      & 95\%     \\ \hline
\end{tabular}%
}
\end{minipage}
\hspace*{\fill}
\vspace{-0.15in}
\end{table}

\subsection{Adaptive Attacks}

When the attacker has the full knowledge of {\name}, one potential adaptive attack that can bypass our method is to force the left and the right terms in Inequation \ref{eq:keyObservation} to have similar values. Based on this intuition, we design an adaptive attack that adds an additional loss term $L_{adp}$ with a weight $\lambda_{adp}$ to the original loss function of the chosen backdoor attack. We define $L_{adp}$ as:\begin{equation}\small
    L_{adp}= \Big|1-\frac{1}{d}\sum_{j=1}^{d}{{
    \sigma^2\left(\big\{\mathcal{B}(F(\mathcal{A}(x_i)), w_j)\big\}_{i=1}^{N_p} \right)}\big/{\sigma^2\left(\big\{\mathcal{B}(F((x_i), w_j)\big\}_{i=1}^{N_b} \right)}}\Big|,
\end{equation}where $N_p$ and $N_b$ are the numbers of poisoned inputs and benign inputs in a minibatch. 
The loss term $L_{adp}$ tries to break the feature-space observation by 
enforcing RAV to be close to one.  We generate ten backdoored DRMs by the adaptive attack with $\lambda_{adp}=0.02$. Table \ref{Tab:adaptive_attack} shows the identification accuracy and the averaged AE and DAE over ten backdoored DRMs.  
The AE of the adaptive attack is significantly higher than that of WaNet. This proves that our feature-space observation of the backdoored DRM is the key characteristic leading to the backdoor behavior. 
The adaptive attack cannot reduce the identification accuracy of our method. 

\section{Discussion and Limitation}

\textbf{Discussion.} To further investigate the performance of {\name}, we evaluate {\name} by considering more backdoor attacks, different architectures of DRMs, and a larger set of DRMs in Appendix \ref{App:more_experiments}. The results show {\name} can consistently defend against various backdoor attacks and can be generalized to different architectures. Also, {\name} maintains a similar identification accuracy on a larger set of DRMs that contains more backdoored and benign DRMs.

\textbf{Limitation.} Similar to backdoor defenses~\citep{wang2019neural,wang2022rethinking} for DCM, our method requires a small benign dataset to identify backdoored DRM and mitigate backdoor behaviors. 

\section{Conclusion}
We propose the first backdoor identification method {\name} for deep regression models in the image domain. Our method fills in the gap where existing backdoor identification methods only focus on deep classification models. Our comprehensive evaluation shows that our method can defend against both input-independent and input-aware backdoor attacks on various datasets.

\bibliography{iclr2024_conference}
\bibliographystyle{iclr2024_conference}

\newpage
\appendix
\section{Appendix}

\myparatight{Roadmap} We provide the details of the momentum reverse trigger in subsection~\ref{App:momentum}. The details of two regression tasks and four datasets are discussed in subsection~\ref{App:datasets}. Subsection~\ref{App:attack} describes the details of backdoor attacks and discusses how to generalize them to regression tasks. Then, we show the details about how to generalize Neural Cleanse~\citep{wang2019neural} and FeatureRE~\citep{wang2022rethinking} to the deep regression model (DRM) in subsection~\ref{App:baseline_defenses}. Next, we visualize the estimation of the target vector during the reverse engineering process and the reversed poisoned images in subsection~\ref{App:target_vector} and subsection~\ref{App:reversed_trigger}, respectively. After that, we detail multi-target backdoor attack for DRMs in subsection~\ref{App:multi_target_attack}. We evaluate {\name} on more backdoor attacks, different architectures of DRMs, and a larger set of DRMs in Appendix \ref{App:more_experiments}. We show that our idea of minimizing the variance can be extended to classification domain for trigger reverse engineering in Appendix \ref{App:DCM}. Finally, we report the technical details of {\name} in subsection~\ref{App:technical_details}.

\myparatight{Experiment environment}We conduct experiments with Python 3.7.13 and Tensorflow 2.9.0 on an Ubuntu 20.04 machine with a NVIDIA A10 GPU. 

\subsection{Details of Momentum Reverse Trigger}
\label{App:momentum}

\begin{figure}[t]
	\centering
	\subfigure[]{\includegraphics[scale=0.3]{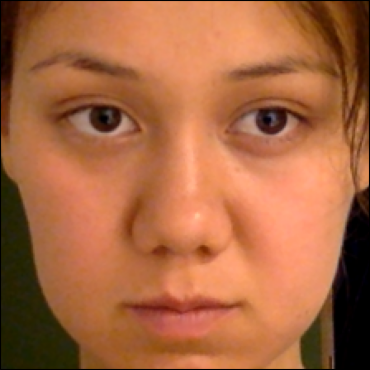}\label{}
}\quad 
	\subfigure[]{\includegraphics[scale=0.3]{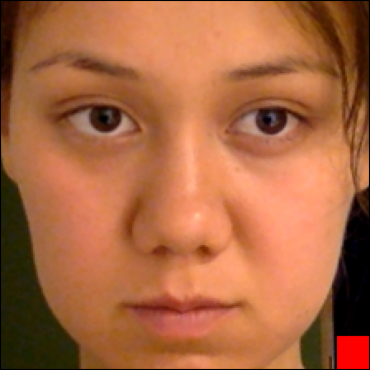}\label{fig:poisoned}
}\quad
 \subfigure[]{\includegraphics[scale=0.3]{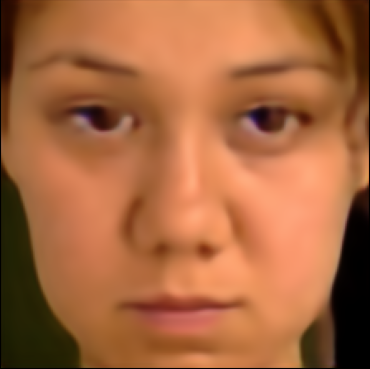}\label{fig:reversed}
} \quad
	\subfigure[]{\includegraphics[scale=0.3]{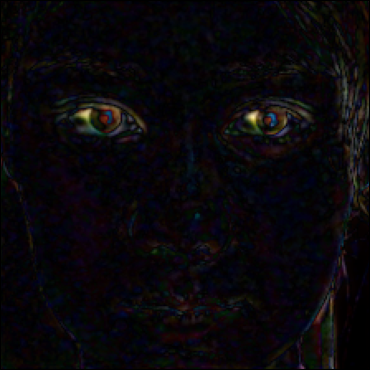}\label{fig:residual}
}
	\caption{The visualization of the sub-optimal solution: 
 (a) the benign image; (b) the poisoned image; (c) the reversed poisoned image when directly solving the optimization problem defined in Equation \ref{opt:REforDRM}; and (d) the 
 residual map between the benign and reversed poisoned images. Solving the optimization problem \ref{opt:REforDRM} fails to reverse the trigger but adds perturbations to the image region that contains the most important features for gaze estimation. 
 }	
	\label{fig:sub_optimal}
\end{figure}

\myparatight{Illustration of the sub-optimal solution} We illustrate the sub-optimal solution of the optimization problem defined in Equation \ref{opt:REforDRM} in Figure \ref{fig:sub_optimal}. Specifically, we use BadNets \citep{badnet} to train a backdoored DRM on MPIIFaceGaze dataset for gaze estimation, where the trigger is a red square, added at the right bottom corner of the input images, as shown in Figure~\ref{fig:poisoned}. Then, we optimize $G_{\theta}$ by solving the optimization problem defined in Equation~\ref{opt:REforDRM}. We show the residual map between the benign image and the reversed poisoned image in Figure \ref{fig:residual}. We can see that solving the optimization problem~\ref{opt:REforDRM} fails to reverse the trigger but adds perturbations to the eyes region that contains the most important features for gaze estimation \citep{Zhang_2017_CVPR_Workshops}. 

\myparatight{Technical details} The idea of momentum reverse trigger is to assign different weights to different regions to balance the attention of the DRM on the image, such that the algorithm can pay attention to all the pixels and search for the trigger that are injected into both important and unimportant regions. To do this, we first generate an \emph{attention map} $\mathcal{T}(x_i) \in \mathbb{R}^{N_w\times N_h}$ for each input $x_i$ based on the gradient of $f$ w.r.t. $x_i$. In detail, for each pixel $x_i[a,b]$, we obtain the corresponding value $\mathcal{T}(x_i)[a,b]$ in the attention map by $\small\mathcal{T}(x_i)[a,b]=\sum_{c=1}^{N_c}{|{\partial f}/{\partial x_i[a,b,c]}|} \label{eq:attention_map}$. Then, we re-scale the attention map to $[0,1)$ by dividing each component of the attention map by a number that is larger than the maximum value in the attention map. Then, instead of directly feeding $G_{\theta}(x_i)$ to $f$, we use the following image as the poisoned image and feed it to the DRM:
\begin{equation}\small
    x'_i = G_{\theta}(x_i)\odot(1 - \mathcal{T}(x_i)) + x_i \odot \mathcal{T}(x_i).
    \label{eq:momentum}
\end{equation}

\subsection{Details of Regression Tasks and Datasets}
\label{App:datasets}
\subsubsection{Details of regression tasks}
We consider two regression tasks, i.e., gaze estimation and head pose estimation. Below, we introduce the details of each regression task.

\textbf{Gaze estimation:} We consider 3D full-face gaze estimation, where the gaze estimation model aims to estimate the 3D gaze direction from the facial image. The 3D gaze direction is represented by a two-dimensional vector, denoting the yaw and pitch angles of the gaze direction. Following the existing works on gaze estimation \citep{Zhang2020ETHXGaze, Zhang_2017_CVPR_Workshops}, we use angular error as the regression error, which is defined as the angle between the estimated and the real gaze directions. 

\textbf{Head pose estimation:} We consider a head pose estimation model that takes a monocular image input and outputs a three-dimensional vector to denote the Eular angle (yaw, pitch, roll) of head pose \citep{gupta2019nose}. We use $L1$ loss as the regression error.

\subsubsection{Details of datasets}

We consider four benchmark datasets in our evaluation and introduce the details of these datasets in the following. We also introduce the training details of benign DRMs on each dataset. 

\textbf{MPIIFaceGaze} \citep{zhang19_pami} is a benchmark dataset for gaze estimation and is collected from 15 subjects in their daily life. Each subject contains 3,000 images under different backgrounds, illumination conditions, and head poses. The image resolution is 224 $\times$ 224. We use Adam optimizer with a learning rate of $0.0001$ to train the model for 10,000 training steps. 

\textbf{ColumbiaGaze} \citep{CAVE_0324} is a gaze estimation dataset collected from 56 subjects. Each subject has 105 facial images. We crop the facial patches from the original images and resize them to 224 $\times$ 224. We use Adam optimizer with an initial learning rate of $0.001$ to train the model for 10,000 training steps. The learning rate is multiplied by 0.1 at 5,000 and 9,000 training steps.

\textbf{Biwi Kinect Dataset} \citep{fanelli_IJCV} is used for head pose estimation. It is collected from 24 subjects, and each subject has 400 to 900 images. Note that we use the \emph{cropped faces of Biwi dataset (RGB images)} released by \citep{Borghi_2017_CVPR}. We resize the image to 112 $\times$ 112. We use Adam optimizer with a learning rate of $0.0001$ to train the model for 5,000 training steps. 

\textbf{Pandora Dataset} \citep{Borghi_2017_CVPR} is a dataset for head pose estimation, which includes more than 250k RGB and depth images. We use the pre-processed dataset, \emph{Cropped faces (RGB images)}, released by the authors, where the facial images are cropped from the original images. The pre-processed dataset has 100 subjects, and contains more than 120,000 images. We use Adam optimizer with an initial learning rate of $0.0001$ to train the model for 10,000 training steps. 
The learning rate is multiplied by 0.1 at 5,000 training steps.

\subsection{Details of Backdoor Attacks}
\label{App:attack}
In this subsection, we describe the details of the backdoor attacks used in our evaluation and how to adapt them to DRMs. We also show the effectiveness of backdoor attacks on DRMs.

\textbf{BadNets}~\citep{badnet} uses a fixed pattern as the backdoor trigger, and the poisoned inputs are generated by pasting the backdoor trigger on the inputs. In our experiments, we use a 20 $\times$ 20 red patch located at the right-bottom corner as the backdoor trigger. The poisoning rate is $5\%$. 

\textbf{Clean Label}~\citep{turner2019label} also uses a fixed pattern as the backdoor trigger. To be more stealthy, the trigger is only applied to images that belong to the target class for classification tasks. In our experiments, we use a 20 $\times$ 20 red patch located at the right-bottom corner as the backdoor trigger. 
To generalize Clean Label to regression tasks, we apply the trigger to the inputs whose annotations are "close" to the target vector. Formally, we consider the images whose annotations $y$ satisfy $\|y-y_T\|\leq \delta$ as the target group to be poisoned. We apply the PGD attack to half of the images in the target group to generate adversarial samples. To improve the performance of backdoor attacks, we apply the trigger to both the adversarial samples and another half of the images in the target group. This is because if we only apply trigger to the adversarial samples, the trained DRM will regard the perturbations generated by the PGD attack as the backdoor trigger and ignore the attacker-defined trigger pattern. Annotations of poisoned images are changed to the target vector.
    

\textbf{WaNet}~\citep{nguyen2021wanet} generates stealth backdoor triggers through image warping techniques. The trigger is inserted into the images by applying the elastic warping operation. Note that, WaNet needs to modify the standard training process to train the backdoored DRM, while BadNets and Clean Label follow the standard way to train the backdoored DRM. To improve the performance of the attack, we set the warping strength to 1. We set the grid sizes for gaze estimation datasets and head pose estimation datasets to 28 and 14, respectively.
    
\textbf{Input-aware dynamic attack (IA)} ~\citep{nguyen2020input} generates dynamic backdoor triggers by using a trainable trigger generator, which takes benign images as inputs and outputs triggers varying from input to input. A trigger generated by an input image cannot be used on another one. Similar to WaNet, IA also needs to modify the training process.

\begin{table}[h]{}
\scriptsize
\centering
\caption{Effectiveness of different backdoor attacks on DRMs.}
\label{Tab:effectiveness_attack}
\resizebox{0.65\linewidth}{!}{%
\begin{tabular}{cccccc}
\\[-2ex]
\hline
\\[-2ex]
Metric & BadNets & IA & Clean Label & WaNet & Benign \\ \hline\\[-2ex]
AE & 3.25 & 3.19  & 0.72  & 1.31 & n/a\ \\
RE & 2.61 & 2.33  & 1.82  & 2.32 & 2.35\\ \hline
\end{tabular}%
}
\end{table}

\myparatight{Effectiveness of backdoor attacks on DRMs} We use attack error (AE) to evaluate the effectiveness of backdoor attacks on DRMs. Given a set of poisoned inputs, we define AE as the average regression error calculated from the output vectors and the target vector over all the poisoned inputs. AE can be regarded as the counterpart to the attack success rate for backdoor attacks on deep classification models. Besides, we define RE as the average regression error calculated from the output vectors and the correct annotations over a benign dataset. For each attack, we apply the corresponding trigger function to each image in the benign dataset $\mathcal{D}_{be}$ to generate a set of poisoned images $\mathcal{PD}_{te}$. We report AE over $\mathcal{PD}_{te}$ and RE over $\mathcal{D}_{be}$ to show the effectiveness of the above mentioned backdoor attacks on DRMs in Table \ref{Tab:effectiveness_attack}. The experimental results show that the AE of different backdoored DRMs are pretty low, which are almost similar to the RE of these backdoored DRMs. Also, the RE of these backdoored DRMs are similar to that of the benign DRM. This proves the effectiveness of these backdoor attacks on DRMs.

\subsection{Details of Baseline Defenses}
\label{App:baseline_defenses}
We generalize two state-of-the-art defenses, i.e., Neural Cleanse (NC) \citep{wang2019neural} and FeatureRE \citep{wang2022rethinking}, to regression models. Since the output of DRM is defined in the continuous space, we generalize them from classification models to regression models by taking the potential target vector $y_t$ as the optimization variable. Below, we introduce the formal definitions of the optimization problems for these two defenses after generalization.

\myparatight{Generalization of NC} We generalize the optimization problem defined in NC to DRMs as:
\begin{equation}
    m^*, \Delta^*, y_t^* = \min_{m,\Delta,y_t}{\frac{1}{N} \sum_{i=1}^N{\ell(y_t,f(A(x_i,m,\Delta))}+\lambda \cdot |m|},
\end{equation}where $\lambda$ is the weight for the second objective; $A(\cdot)$ is a function that applies a trigger represented by $m$ and $\Delta$ to the benign image $x$; and $\ell(\cdot)$ is the loss function for training the DRMs. Specifically, $\ell(\cdot)$ is $\ell_1$ loss for gaze estimation and head pose estimation. The detailed description of $A(\cdot)$ can be found in \citep{wang2019neural}.

\myparatight{Generalization of FeatureRE} We generalize the optimization problem defined in FeatureRE to DRMs as:
\begin{align}
\begin{aligned}
    \theta^*, m^*, y_t^*=&\min_{\theta, m, y_t}\ell(H((1-m)\odot a+m\odot t), y_t)+\frac{\lambda_3}{N} \sum_{i=1}^N{\|G_{\theta}(x_i)-x_i\|},\\
    \text{where}& \;\; t=mean(m\odot F(G_{\theta}(\{x_i\}_{i=1}^N))), a=F(\{x_i\}_{i=1}^N),\\
    &\text{s.t.}\;\; std(m\odot F(G_{\theta}(\{x_i\}_{i=1}^N)))\leq \tau_1, \|m\|\leq \tau_2,
    \end{aligned}
\end{align}where $m$ is the feature-space mask; $\lambda_3$ is the weight for the second objective; $\tau_1$ and $\tau_2$ are the thresholds for two constraints, respectively. 

\subsection{Estimation of the Target Vector}
\label{App:target_vector}
\begin{figure}[]
	\centering
	\includegraphics[width=13.5cm]{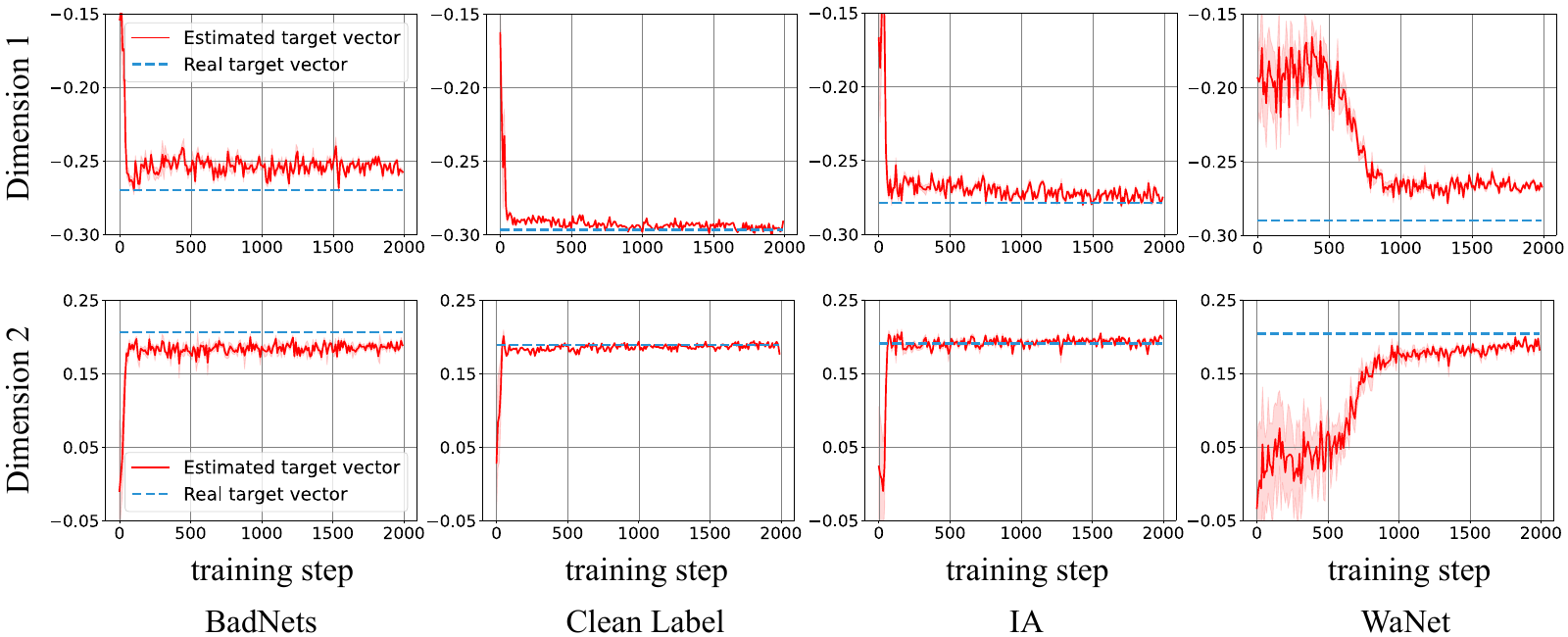}
	\caption{Visualization of the estimation of the target vector (two-dimensional vector) for DRMs backdoored by different attacks on MPIIFaceGaze dataset during the reverse engineering process. The two rows correspond to the first and second dimensions of the output of the DRMs. The red curves denote the estimation of the corresponding dimension of the target vector, while the blue curves denote that of the real target vector. The red curves can converge to the neighbor of the blue curves, which means that {\name} can estimate the target vector.}
	\label{fig:target_curve}

\end{figure}

 {\name} is able to estimate the target vector. Given a set of reversed poisoned images, we use the mean vector of the corresponding output vectors as the estimation of the target vector. We visualize the estimation of the target vector during the reverse engineering process for different backdoor attacks, i.e., BadNets, Clean Label, IA, and WaNet, on MPIIFaceGaze dataset in Figure~\ref{fig:target_curve}. Note that, since it is infeasible to train a backdoored DRM $f$ that precisely outputs the attacker-chosen target vector for poisoned images, we use the vector, $\mathbb{E}_{x\in \mathcal{D}_{be}}{f(\mathcal{A}(x))}$, as the \emph{real target vector}, where $\mathbb{E}(\cdot)$ is a function to obtain the mean vector for a given set of vectors. Figure~\ref{fig:target_curve} shows that the output vectors of the reversed poisoned images can converge to the neighbor of the real target vector.
 
\subsection{Visualization of Reversed Poisoned Images}
\label{App:reversed_trigger}

\begin{figure}[]
	\centering
	\subfigure[Benign images.]{\includegraphics[scale=0.46]{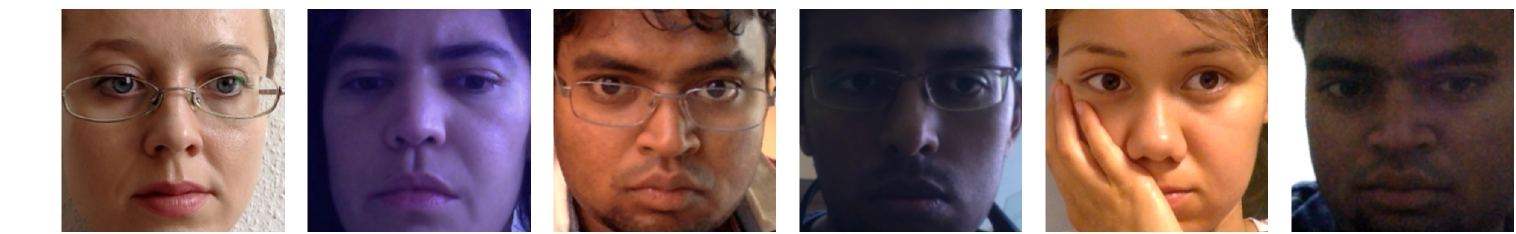}
 \label{fig:benign_images}} 
	\subfigure[BadNets.]{\includegraphics[scale=0.46]{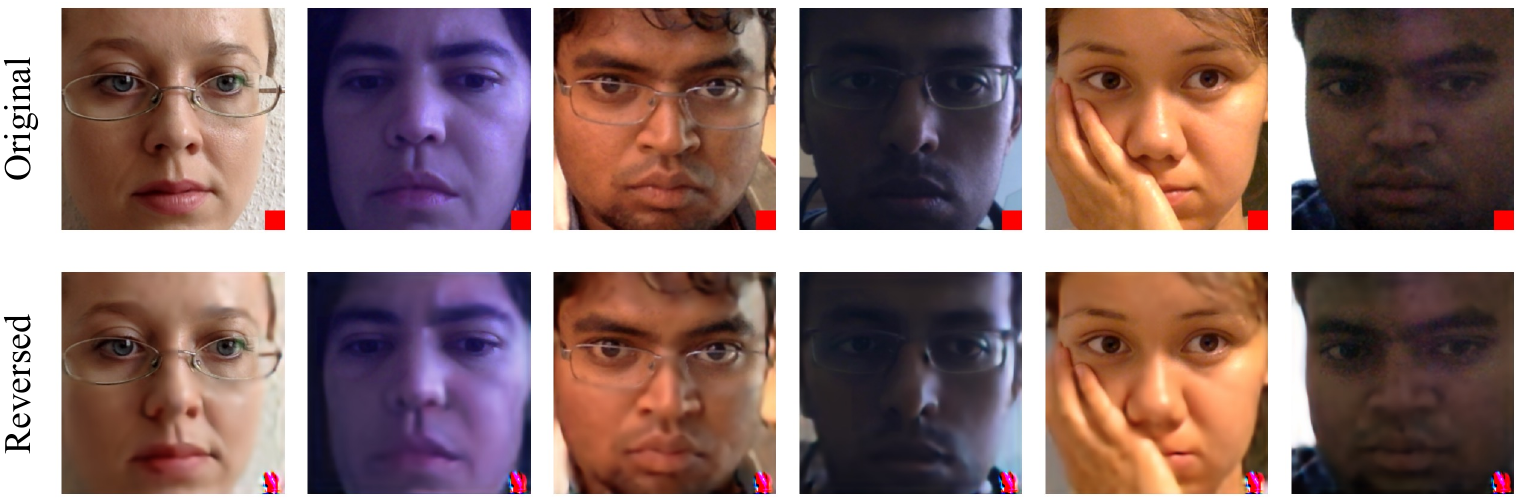}\label{fig:badnet_images}} 
 \subfigure[Clean Label.]{\includegraphics[scale=0.46]{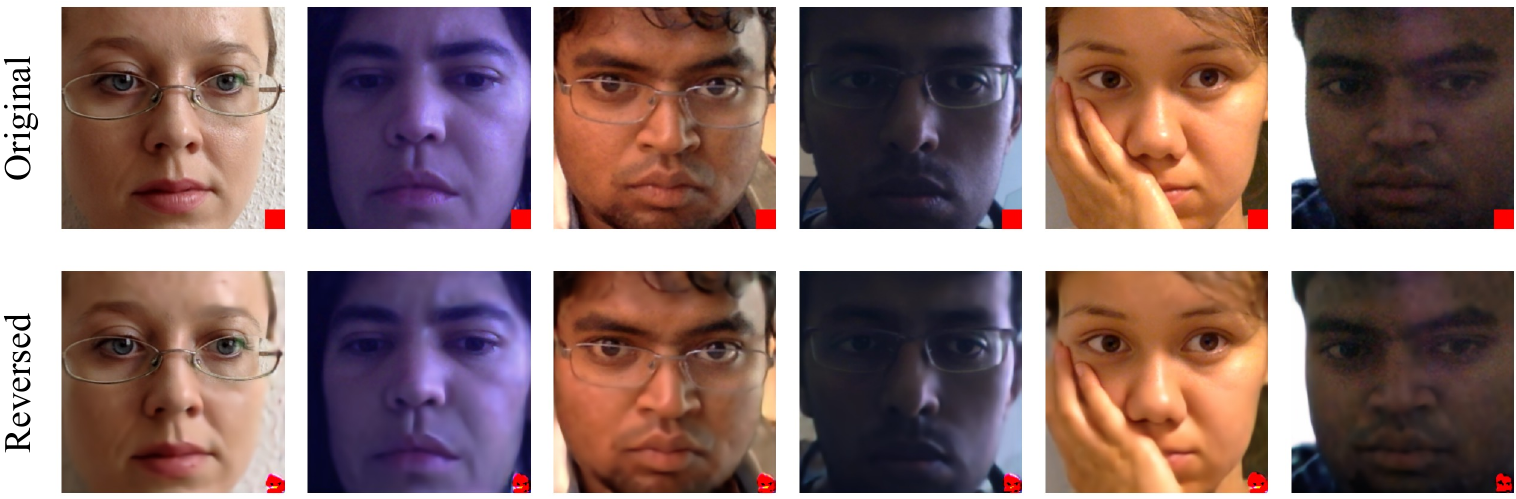}\label{fig:cl_images}}
 \subfigure[IA.]{\includegraphics[scale=0.46]{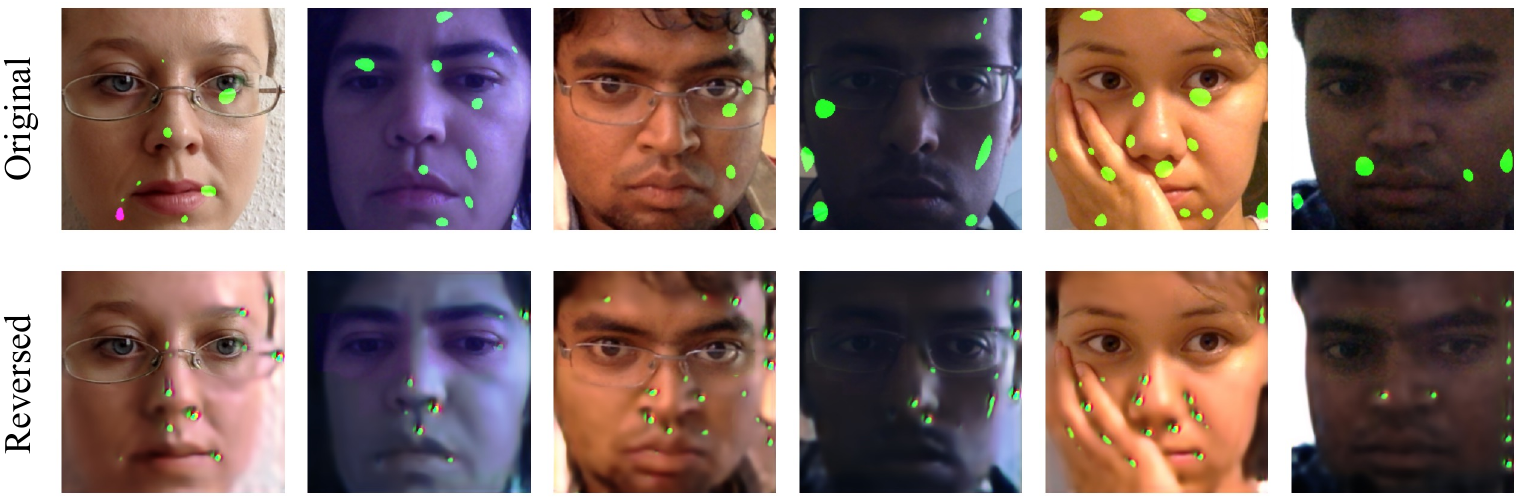}\label{fig:IA_images}}
 \subfigure[WaNet.]{\includegraphics[scale=0.46]{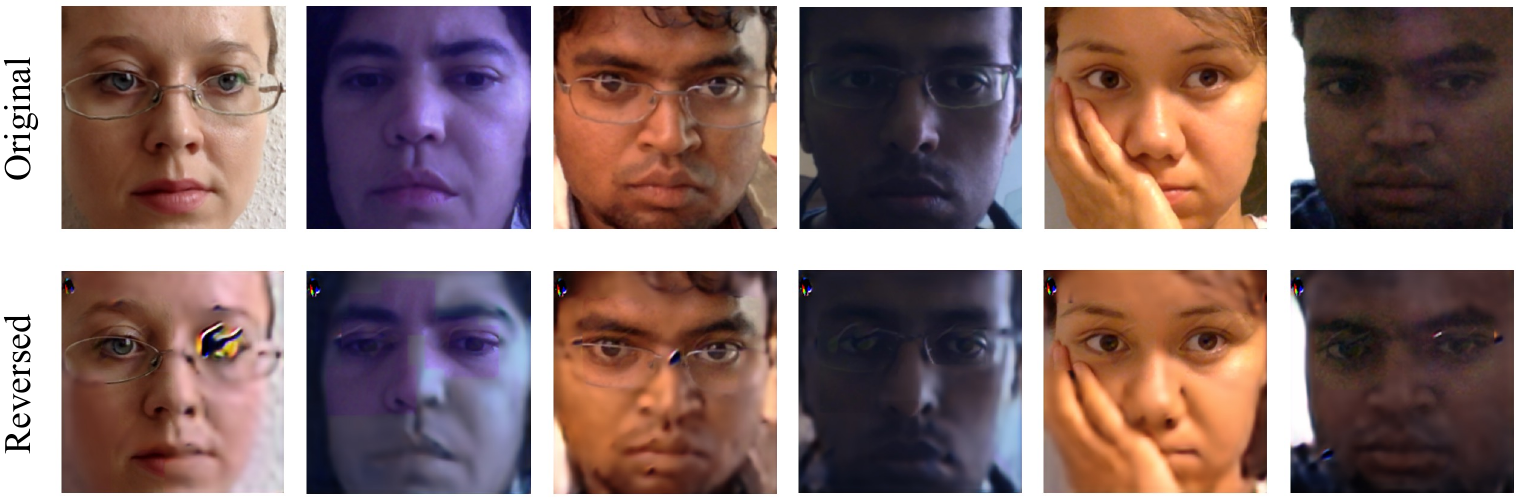}\label{fig:wanet_images}}

\caption{Comparison between (a) the benign images, and the original poisoned images and the corresponding reversed poisoned images for (b) BadNets, (c) Clean Label, (d) IA, and (e) WaNet. The reversed poisoned images are close to the original poisoned images.
}	
	\label{fig:reversed_trigger}

\end{figure}

To investigate if {\name} can reverse engineer the poisoned images, we show the benign images, the original poisoned images generated by different backdoor attacks, and the reversed poisoned images in Figure~\ref{fig:reversed_trigger}. Specifically, we randomly sample six benign images from the benign dataset $\mathcal{D}_{be}$ and show them in Figure~\ref{fig:benign_images}. We then show the original poisoned images generated by BadNets, Clean Label, IA, and WaNet, on MPIIFaceGaze dataset and the corresponding reversed poisoned images in Figure~\ref{fig:badnet_images}, Figure~\ref{fig:cl_images}, Figure~\ref{fig:IA_images}, and Figure~\ref{fig:wanet_images}, respectively. {\name} is able to reverse engineer the poisoned images that are close to the original poisoned images.

\subsection{Multi-target Attack}
\label{App:multi_target_attack}

\begin{wraptable}[6]{}{4.5cm}
\scriptsize
\vspace{-13pt}
\centering
\caption{Backdoor identification performance of {\name} for multi-target WaNet.}
\label{Tab:multi_target}
\resizebox{0.95\linewidth}{!}{%
\begin{tabular}{ccccc}
\hline
\\[-2ex]
 TP & FP & FN & TN & Acc  \\ \hline\\[-2ex]
10 & 1 & 0  & 9  & 95\%\\ \hline
\end{tabular}%
}
\end{wraptable}

We consider the scenario that the DRM is backdoored by multiple trigger functions with different target vectors. We generalize WaNet such that it can insert two warping-based triggers into a DRM. Specifically, we generate two different warping functions, and each of them corresponds to a unique target vector. During the training process, we randomly select $15\%$ images in the minibatch to apply the first warping function and change their annotations to the corresponding target vector. We then select another $15\%$ images in the minibatch to apply the second warping function and change their annotations to the corresponding target vector. We do not modify other procedures of WaNet. We call the generalized WaNet \emph{multi-target WaNet}. We train ten backdoored DRMs by using multi-target WaNet and ten benign DRMs on MPIIFaceGaze dataset. We report the performance of backdoor identification of {\name} in this scenario in Table~\ref{Tab:multi_target}, which shows that our method is effective on identifying DRMs with multiple backdoors.

\subsection{More Experiments}
\label{App:more_experiments}

\myparatight{Evaluation on more architectures} 
We study the generalization ability of {\name} on different architectures of DRMs. Specifically, beyond ResNet18, we consider two different architectures, i.e., ResNet34 \citep{He_2016_CVPR} and MobileNetV1 \citep{howard2017mobilenets}. For each architecture, we train ten backdoored DRMs by WaNet and ten benign DRMs on MPIIFaceGaze dataset. We report backdoor identification performance and backdoor mitigation performance for different architectures in Table \ref{Tab_idenfication_Architectures} and Table \ref{Tab_mitigation_Architectures}, respectively. The results show that {\name} can be generalized to different architectures of DRMs.

\begin{table}[]
\vspace{-10pt}
\begin{minipage}[c]{0.48\textwidth}
\centering
\caption{Backdoor identification performance of {\name} for different architectures. 
}
\resizebox{.9\linewidth}{!}{
\begin{tabular}{cccccc}
\\[-2ex]
\hline
\\[-2ex]
Architecture       & TP & FP & FN & TN & Acc   \\ \hline\\[-1.5ex]
ResNet18    & 10 & 1  & 0  & 9  & 95\%  \\
ResNet34    & 10 & 0  & 0  & 10 & 100\% \\
MobileNetV1 & 7  & 0  & 3  & 10 & 85\%  \\ \hline
\end{tabular}}
\label{Tab_idenfication_Architectures}
\end{minipage}
\quad
\hfill
\begin{minipage}[c]{0.48\textwidth}
\centering
\caption{Backdoor mitigation performance of {\name} for different architectures. 
}
\resizebox{.9\linewidth}{!}{
\begin{tabular}{ccccc}
\\[-2ex]
\hline
\\[-2ex]
\multirow{2}{*}{Architecture} & \multicolumn{2}{c}{Undefended} & \multicolumn{2}{c}{\name} \\ \cline{2-5} \\[-1.5ex]
                        & AE             & DAE           & AE           & DAE           \\ \hline\\[-1.5ex]
ResNet18                 & 1.31           & 15.90         & 15.36        & 3.29         \\ 
ResNet34                      & 2.82           & 15.97          & 15.50          & 2.40         \\ 
MobileNetV1                   & 0.42           & 16.04          & 15.88        & 3.70        \\ \hline

\end{tabular}}
\label{Tab_mitigation_Architectures}
\end{minipage}
\vspace{-3pt}
\end{table}

\myparatight{Evaluation on more attacks} We further evaluate the performance of {\name} on defending Blend attack \citep{8802997} and SIG~\citep{liu2020reflection}. For each attack, we train ten backdoored DRMs by WaNet and ten benign DRMs on MPIIFaceGaze dataset. We report backdoor identification performance in Table \ref{Tab_idenfication_more_attack}. We also report backdoor mitigation performance in Table \ref{Tab_mitigation_more_attack}. The experimental results show that {\name} can consistently defend against various attacks.

\begin{table}[]
\begin{minipage}[c]{0.48\textwidth}
\centering
\caption{Backdoor identification performance of {\name} on more attacks. 
}
\resizebox{.8\linewidth}{!}{
\begin{tabular}{cccccc}
\\[-2ex]
\hline
\\[-2ex]
Attack       & TP & FP & FN & TN & Acc   \\ \hline\\[-1.5ex]
Blend        & 10 & 1  & 0  & 9  & 95\%  \\
SIG          & 10 & 1  & 0  & 9  & 95\%  \\ \hline
\end{tabular}}
\label{Tab_idenfication_more_attack}
\end{minipage}
\quad
\hfill
\begin{minipage}[c]{0.48\textwidth}
\centering
\caption{Backdoor mitigation performance of {\name} on more attacks. 
}
\resizebox{.83\linewidth}{!}{
\begin{tabular}{ccccc}
\\[-2ex]
\hline
\\[-2ex]
\multirow{2}{*}{Attacks} & \multicolumn{2}{c}{Undefended} & \multicolumn{2}{c}{\name} \\ \cline{2-5} \\[-2ex]
                        & AE             & DAE           & AE           & DAE           \\ \hline\\[-2ex]
Blend                   & 2.94           & 14.51         & 12.82        & 4.56         \\ 
SIG                     & 1.83           & 15.96         & 16.11        & 2.93        \\ \hline

\end{tabular}}
\label{Tab_mitigation_more_attack}
\end{minipage}
\vspace{-10pt}
\end{table}

\begin{wraptable}[]{}{5cm}
\vspace{-13pt}
\scriptsize
\centering
\caption{Backdoor identification performance of {\name} on a larger set of DRMs.}
\label{Tab_more_DRMs}
\resizebox{0.875\linewidth}{!}{%
\begin{tabular}{ccccc}
\hline
\\[-1.5ex]
 TP & FP & FN & TN & Acc  \\ \hline\\[-1.5ex]
29 & 5 & 1  & 25  & 90\%\\ \hline
\end{tabular}%
}

\end{wraptable}

\myparatight{Evaluation on a larger set of DRMs} We evaluate the backdoor identification performance of {\name} on a larger set of DRMs. Specifically, we train 30 backdoored DRMs by WaNet and 30 benign DRMs on MPIIFaceGaze dataset. We report backdoor identification performance in Table \ref{Tab_more_DRMs}. The results show {\name} can reach the identification accuracy at $90\%$ on a larger set of DRMs.

\subsection{Generalization to classification domain}
\label{App:DCM}
We show that the idea of \emph{minimizing the variance in the output space} can be extended to DCMs. By doing so, we can identify backdoored DCMs without enumerating all the labels. 

We use $\{S(C(A(x_i)))\}_{i=1}^N$ to denote the probability vectors obtained by a backdoored DCM $C(\cdot)$ from a set of poisoned images $\{\mathcal{A}(x_i)\}_{i=1}^N$, where $S(\cdot)$ is the softmax function. Intuitively, the poisoned images will lead to similar probability vectors for a backdoored DCM. Therefore, $\frac{1}{d}\sum_{j=1}^d{\sigma^2(\{S_j(C(A(x_i)))\}_{i=1}^N})$ will be a small positive value, where $S_j(C(A(x_i)))$ is the $j$th component of $S(C(A(x_i)))$. Based on this intuition, we propose the following optimization objective to generalize DRMGuard from the regression domain to the classification domain.
\begin{equation}
    \theta^{*}=\min_{\theta} \frac{\lambda_1}{d}\sum_{j=1}^d{\sigma^2\left(\{S_j(f\left(G_{\theta}(x_i)\right))\}_{i=1}^N\right)} + \frac{1}{N} \sum_{i=1}^N\|G_{\theta}(x_i)-x_i\|_1
\end{equation}

\begin{wraptable}[5]{}{5cm}
\vspace{-13pt}
\scriptsize
\centering
\caption{Backdoor identification performance for DCMs.}
\label{Tab_DCMs}
\resizebox{0.875\linewidth}{!}{%
\begin{tabular}{ccccc}
\hline
\\[-1.5ex]
 TP & FP & FN & TN & Acc  \\ \hline\\[-1.5ex]
7 & 0 & 3  & 10  & 85\%\\ \hline
\end{tabular}%
}

\end{wraptable}
The first objective in the optimization problem aims to reverse engineer the poisoned images $\{G_{\theta}(x_i)\}_{i=1}^N$ that lead to the similar probability vectors, regardless of their actual contents. The second optimization term ensures the transformed images $G_{\theta}(x_i)$ is similar to original image $x_i$. We remove the feature-space regularization term designed for DRMs, since backdoored DRMs and backdoored DCMs have different feature-space characteristics. 

\begin{figure}[]
	\centering
	\includegraphics[width=8.5cm]{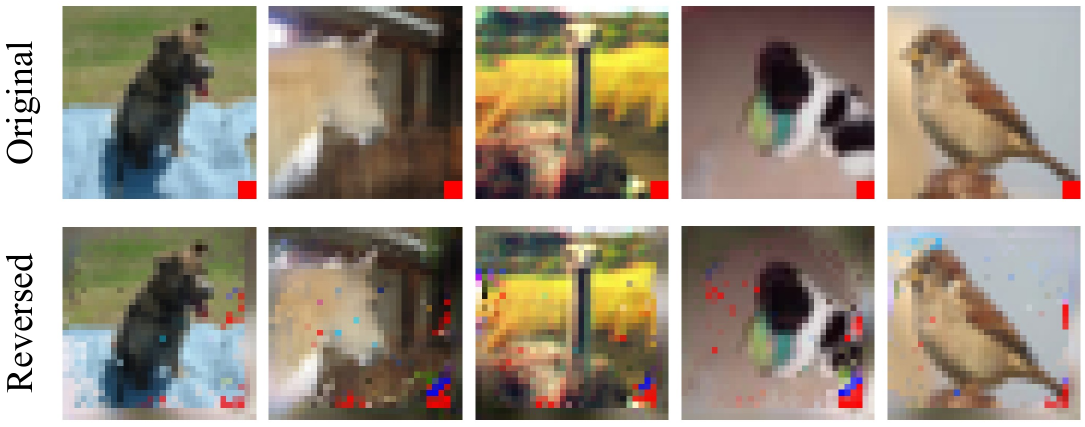}
	\caption{Illustration of original poisoned images (the first row) and reversed poisoned image (the second row) for backdoored DCMs,}
	\label{fig:piosined_DCM}

\end{figure}

We train ten backdoored DCMs by BadNets and ten benign DCMs on Cifar10.  We report backdoor identification results in the Table \ref{Tab_DCMs}. We visualize the reversed poisoned image in Figure \ref{fig:piosined_DCM}. The results show that our idea of minimizing the variance in the output space can be extended to the classification domain for backdoor identification without enumerating all the labels.

\subsection{Technical Details of {\name}}
\label{App:technical_details}
We use a simple generative model to implement $G_{\theta}$, which is similar to the generative model used in~\citep{nguyen2020input}. Before performing the reverse engineering, we pre-train $G_{\theta}$ on the benign dataset for 5,000 training steps. During the reverse engineering, the batch size for gaze estimation datasets is 50, while that for head pose estimation datasets is 100. 
We use Adam optimizer to train $G_{\theta}$ for 2,000 training steps, taking about 16 minutes. The learning rates for gaze estimation and head pose estimation are 0.0015 and 0.0001, respectively.

\newpage
\end{document}